%% file: main.tex
\documentclass[10pt,twocolumn,letterpaper]{article}
\usepackage{cvpr}              

\usepackage{graphicx}
\usepackage{amsmath}
\usepackage{amssymb}
\usepackage{booktabs}

\input{includesAndMacros_yagiz}

\usepackage[pagebackref,breaklinks,colorlinks]{hyperref}
\usepackage{xcolor}

\usepackage[capitalize]{cleveref}
\crefname{section}{Sec.}{Secs.}
\Crefname{section}{Section}{Sections}
\Crefname{table}{Table}{Tables}
\crefname{table}{Tab.}{Tabs.}

\def\cvprPaperID{7040} 
\def\confName{CVPR}
\def\confYear{2023}

\begin{document}

\placetextbox{0.13}{0.02}{\includegraphics[width=4cm]{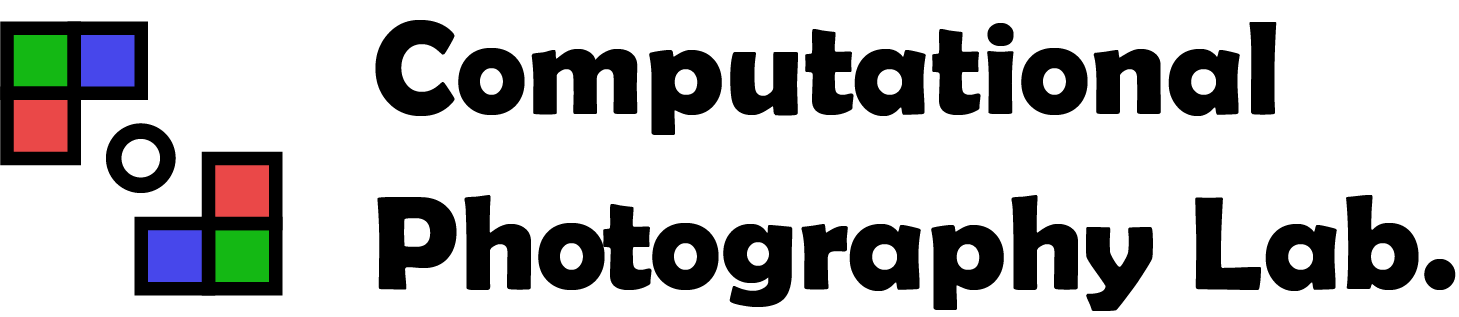}}
\placetextbox{0.78}{0.02}{Find the project web page here:}
\placetextbox{0.78}{0.035}{\textcolor{purple}{\url{http://yaksoy.github.io/realisticEditing/}}}

\title{Realistic Saliency Guided Image Enhancement\negvspace}

\author{
S. Mahdi H. Miangoleh$^{1}$\quad 
Zoya Bylinskii$^{2}$\quad
Eric Kee$^{2}$\quad
Eli Shechtman$^{2}$\quad
Ya\u{g}{\i}z Aksoy$^{1}$
\negvspace\\\\
$^{1}$ Simon Fraser University \quad\quad $^{2}$ Adobe Research
}

\twocolumn[{%
\renewcommand\twocolumn[1][]{#1}%
\maketitle
\begin{center}
    \centering
    \showimagew[\linewidth]{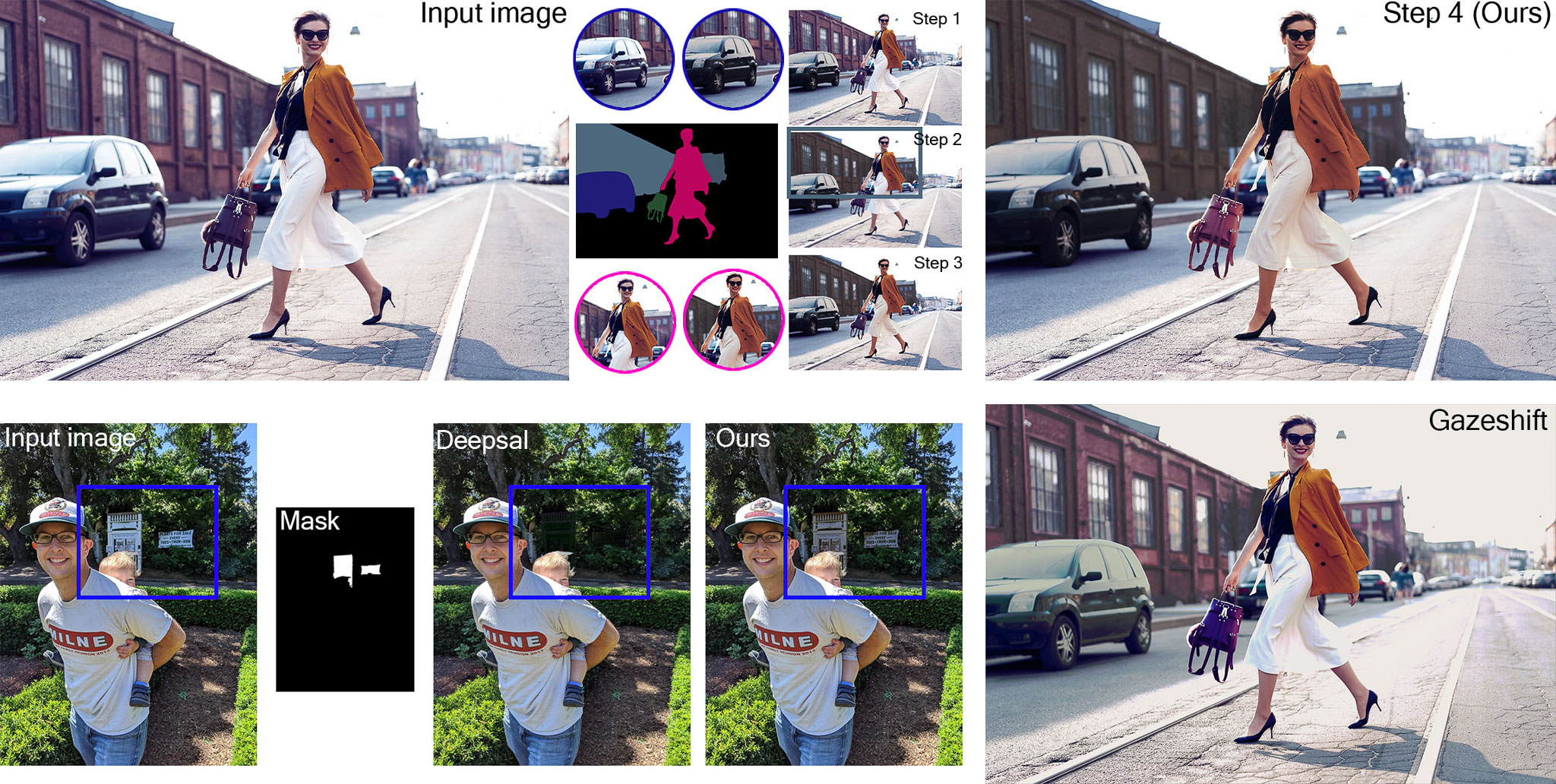}
    \vspace{-0.75cm}
    \captionof{figure}{
    (top) We develop a saliency-based image enhancement method that can be applied to multiple regions in the image to de-emphasize objects (steps 1, 2) or enhance subjects (steps 3, 4). (bottom) Our novel realism loss allows us to apply realistic edits to a wide variety of objects while state-of-the-art methods~\cite{aberman2021deep,gazeshift} may generate less realistic editing results. 
    }
    \label{fig:teaser}
\end{center}%
}]

\begin{abstract}
\input{tex/0-abstract}
\end{abstract}

\vspace{-1.2cm}
\section{Introduction}

\input{tex/1-introduction}
\label{sec:intro}

\section{Related Work}
\input{tex/2-relatedwork}

\label{sec:relatedwork}

\section{Realism Network}
\input{tex/3-realismNetwork}

\label{sec:realismmetric}

\section{Saliency Guided Image Enhancement}
\input{tex/4-method}
\label{sec:method}

\section{Experiments and Results}
\input{tex/5-experiments}

\label{sec:experiments}

\section{Conclusion and Future work}
\input{tex/6-conclusion}
\label{sec:conclusion}


{\small
\bibliographystyle{ieee_fullname}
\bibliography{references}
}

\end{document}

%% file: includesAndMacros_yagiz.tex
\usepackage{atbegshi}
\newcommand{\placetextbox}[3]{
  \setbox0=\hbox{#3}
  \AtBeginShipoutNext{\AtBeginShipoutUpperLeft{%
    \put(\dimexpr#1\paperwidth\relax,-\dimexpr#2\paperheight\relax)
    {\vtop{{\null}\makebox[0pt][c]{#3}}}%
  }}%
}

\usepackage[normalem]{ulem}
\usepackage{graphics}
\usepackage{array}
\usepackage{multirow}
\usepackage{wrapfig}
\usepackage{hhline}
\usepackage{pifont}
\newcommand{\showimagew}[2][\linewidth]{\includegraphics[width={#1}]{{#2}}}

\newcolumntype{K}[1]{>{\centering\arraybackslash}p{#1}}
\newcommand{\negvspace}{\vspace{-0.2cm}}

\usepackage{duckuments}
\usepackage{tikzducks}
\usepackage{graphicx}
\usepackage{adjustbox}
\usepackage{pifont}
\definecolor{islamicgreen}{rgb}{0.0, 0.56, 0.0}

\newcommand{\myhref}[3][blue]{\href{#2}{\color{#1}{#3}}}
\usepackage{float}

\newcommand{\ts}{\textsuperscript}

%% file: tex/0-abstract.tex
\negvspace
\negvspace

Common editing operations performed by professional photographers include the cleanup operations: de-emphasizing distracting elements and enhancing subjects. 
These edits are challenging, requiring a delicate balance between manipulating the viewer's attention while maintaining photo realism. 
While recent approaches can boast successful examples of attention attenuation or amplification, most of them also suffer from frequent unrealistic edits. 
We propose a realism loss for saliency-guided image enhancement to maintain high realism across varying image types, while attenuating distractors and amplifying objects of interest. 
Evaluations with professional photographers confirm that we achieve the dual objective of realism and effectiveness, and outperform the recent approaches on their own datasets, while requiring a smaller memory footprint and runtime. We thus offer a viable solution for automating image enhancement and photo cleanup operations. 

\negvspace
\negvspace

%% file: tex/1-introduction.tex
In everyday photography, the composition of a photo typically encompasses subjects on which the photographer intends to focus our attention, rather than other distracting things. When distracting things cannot be avoided, photographers routinely edit their photos to de-emphasize them. Conversely, when the subjects are not sufficiently visible, photographers routinely emphasize them. Among the most common emphasis and de-emphasis operations performed by professionals are the elementary ones: changing the saturation, exposure, or the color of each element. Although conceptually simple, these operations are challenging to apply because they must delicately balance the effects on the viewer attention with photo realism. 

To automate this editing process, recent works use saliency models as a guide~\cite{aberman2021deep, gazeshift, mechrez2019saliency, chen2019guide, Gatys2017human, Jiang_2021_CVPR}.
These saliency models ~\cite{JIA20EML, Fosco_2020_CVPR, margoinPatch13, Pan_2017_SalGAN, kummerer2017understanding} aim to predict the regions in the image that catch the viewer's attention, and saliency-guided image editing methods are optimized to increase or decrease the predicted saliency of a selected region. Optimizing solely based on the predicted saliency, however, often results in unrealistic edits, as illustrated in Fig.~\ref{fig:teaser}. This issue results from the instability of saliency models under the image editing operations, as saliency models are trained on unedited images. Unrealistic edits can have low predicted saliency even when they are highly noticeable to human observers, or vice versa. This was also noted by Aberman et al.~\cite{aberman2021deep}, and is illustrated in Fig.~\ref{fig:salmodel}.

Previous methods tried to enforce realism using adversarial setups~\cite{gazeshift, Jiang_2021_CVPR, chen2019guide, Gatys2017human}, GAN priors~\cite{aberman2021deep, Jiang_2021_CVPR}, or cycle consistency~\cite{chen2019guide} but with limited success (Fig.~\ref{fig:teaser}). Finding the exact point when an image edit stops looking realistic is challenging. Rather than focusing on the entire image, in this work, we propose a method for measuring the realism of a local edit. To train our network, we generate realistic image edits by subtle perturbations to exposure, saturation, color or white balance, as well as very unrealistic edits by applying extreme adjustments. Although our network is trained with only positive and negative examples at the extremes, we successfully learn a continuous measure of realism for a variety of editing operations as shown in Fig.~\ref{fig:realismnetperformance}.

We apply our realism metric to saliency-guided image editing by training the system to optimize the saliency of a selected region while being penalized for deviations from realism. We show that a combined loss allows us to enhance or suppress a selected region successfully while maintaining high realism. Our method can be also be applied to multiple regions in a photograph as shown in Fig.~\ref{fig:teaser}.

Evaluations with professional photographers and photo editors confirm our claim that we maintain high realism and succeed at redirecting attention in the edited photo. Further, our results are robust to different types of images including human faces, and are stable across different permutations of edit parameters. Taken together with our model size of 26Mb and run-time of 8ms, these results demonstrate that we have a more viable solution for broader use than the approaches that are available for these tasks to date.

%% file: tex/2-relatedwork.tex
Various image enhancement methods have been introduced in the literature to amplify a region of interest or de-emphasise distracting regions, improve image aesthetics, and redirect the viewer's attention. This task has been referred to as \emph{attention retargeting}~\cite{Mateescu2014attention} or \emph{re-attentionizing}~\cite{Nguyen2013reatten} as well. Earlier methods~\cite{Mateescu2014attention, Hagiwara2011, Sara2005Deemphasis, wong2011saliency, vazquez2017gamut} incorporated prior knowledge of saliency cues (saturation, sharpness, color, gamut, etc.) to guide the editing process to achieve the desired change in saliency. But, relying solely on saliency cues both limits the diversity of generated edits, and creates unrealistic edits due to the lack of semantic constraints. As our experiments show, OHR~\cite{Mateescu2014attention} tends to generate unrealistic color changes that are semantically incorrect, and WRS~\cite{wong2011saliency} is limited to contrast and saturation adjustments with limited effectiveness.  

\input{figures/salmodels}


\input{tables/realfakedatarange}
\input{figures/realismperformance}

Recent works leverage saliency estimation networks~\cite{JIA20EML, Fosco_2020_CVPR, margoinPatch13, Pan_2017_SalGAN, kummerer2017understanding} to optimize for a desired saliency map instead of relying on prior saliency cues. Saliency models are trained to output a heatmap that represents where human gaze would be concentrated in an image. These models are not trained to respond to the realism of the input image. Hence they might predict an inconsistent decrease or increase in the saliency of a region when unrealistic or semantically implausible edits are applied, which would be otherwise jarring to human viewers (Fig.~\ref{fig:salmodel}). Using saliency as the only supervision can result in unrealistic images. 

To prevent unrealistic edits, prior works enforce constraints on the allowable changes, use adversarial training~\cite{gazeshift, Jiang_2021_CVPR, chen2019guide, Gatys2017human} or exploit learned priors from GAN-based models~\cite{aberman2021deep, Jiang_2021_CVPR}. For instance, Mechrez et al.~\cite{mechrez2019saliency} and Aberman et al. (Warping)~\cite{aberman2021deep} constrain the result to match the input content in order to maintain its appearance. Aberman et al. (CNN and Recolorization)~\cite{aberman2021deep} use a regularization term that limits the amount of change an image can undergo to maintain the realism.  Mejjati et al.~\cite{gazeshift} designed a global parametric approach to limit the editing operations to a set of common photographic ones. 
Chen et al.~\cite{chen2019guide} exploit cycle consistency to keep the output within the domain of the input image. Gatys et al.~\cite{Gatys2017human} use a texture loss alongside the VGG perceptual loss as a proxy for realism.

Lalonde et al.\cite{lalonde2007using} argue that humans prefer color consistency within images, regardless of object semantics. They use color statistics to measure realism and use it to recolor the images to match the background in compositing task. Zhu et al.\cite{zhu2015learning} train a network to discriminate between natural images and computer-generated composites and use it as a realism measure for compositing task. Realism is also a crucial factor in GANs, as highlighted by~\cite{realisticgan}.

We present a new method for estimating the realism of a local edit. Combining our realism loss with saliency guidance, we show that we can successfully apply attention attenuation or amplification while keeping the final result realistic without requiring data annotated with realism or bulky GAN priors to estimate realism.

%% file: figures/salmodels.tex
\begin{figure}[t]

\centering
\footnotesize
\resizebox{\linewidth}{!}{
}

\centering
\footnotesize
\resizebox{\linewidth}{!}{
\begin{tabular}
{  
    K{0.25\linewidth}
    K{0.25\linewidth}
    K{0.25\linewidth}
    K{0.25\linewidth}
}
Original image & Saliency map & Edited image & Saliency map
\end{tabular}
}
\vspace{-0.2cm}
\showimagew[\linewidth]{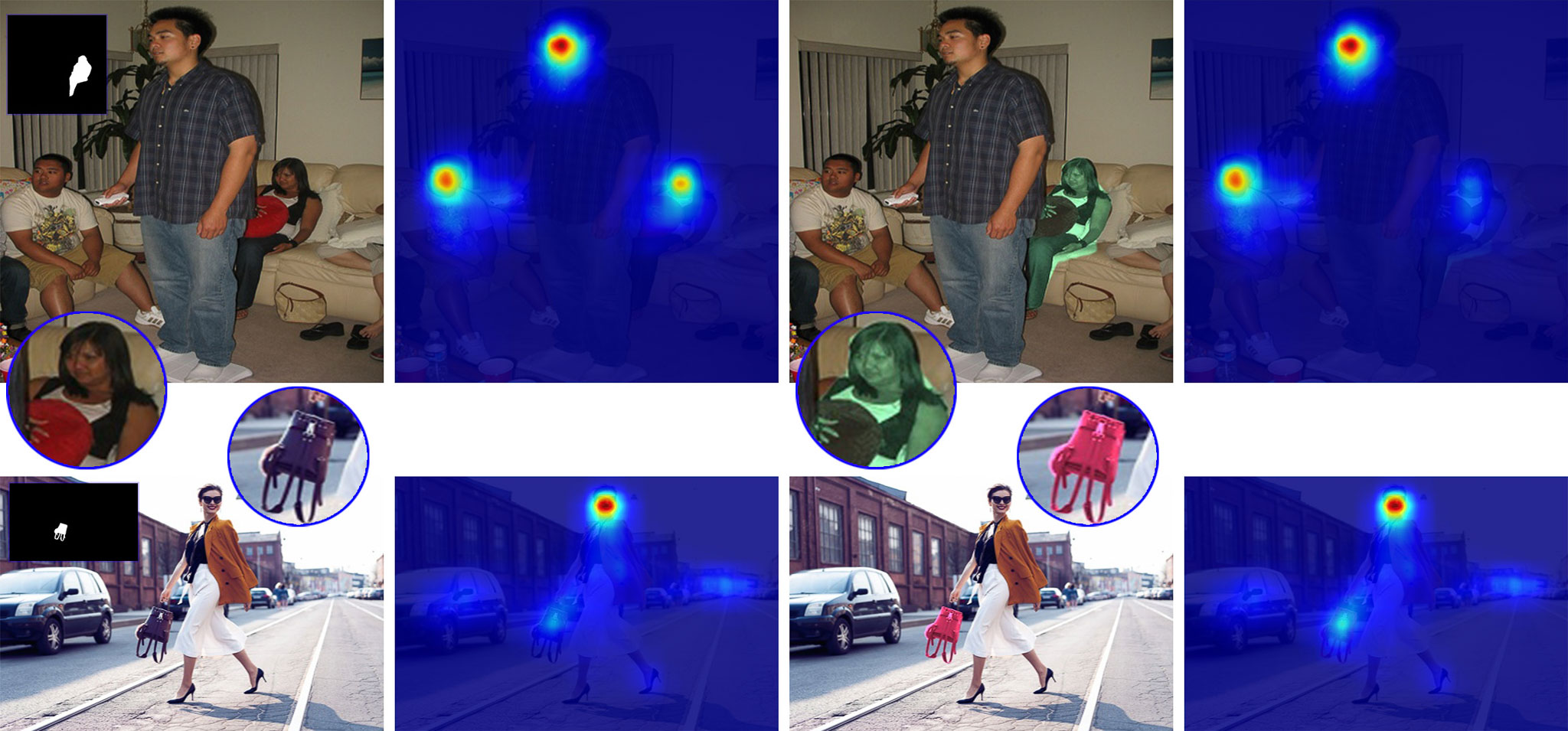}
\caption{
Predicted saliency maps~\cite{JIA20EML} for the original images and edited versions, with extreme edits applied. Note that saliency models are typically trained with realistic images. This makes them susceptible to inaccurate predictions for unrealistic inputs, as the green woman in the top row estimated to have low saliency.}
\vspace{-0.5cm}
\label{fig:salmodel}
\end{figure}

%% file: tables/realfakedatarange.tex
\begin{table*}[t]
\caption{Parameter value ranges used to generate real and fake training images for the realism estimation network.}
\vspace{-0.25cm}
\resizebox{\linewidth}{!}
{
\begin{tabular}{l|ccccc}
              & Exposure                         & Saturation                      & Color curve        & White balancing  & Number of edits    \\
\hline
Real          & $[0.85, 1.15]$                & $[0.85, 1.15]$               & $[0.85, 1.15]$ & Not allowed    & $[1, 3]$ \\
Fake & $[0.5, 0.75] \cup [1.5, 2]$  & $[0, 0.5] \cup [1.5-2]$       & $[0.5, 2]$      & $[0.9, 1]$  &  $[2, 4]$ \\
Fake(human specific)   & $[0.5, 0.75] \cup [1.25, 1.5]$ & $[0.5, 0.75] \cup [1.25, 1.5]$ & $[0.5, 2]$     & Not allowed    & $[2, 3]$ \\

\end{tabular}
}
\label{tab:editranges}
\end{table*}

%% file: figures/realismperformance.tex
\begin{figure*}[t]
    \centering
    \resizebox{\linewidth}{!}
    {
    \includegraphics[width=\textwidth]{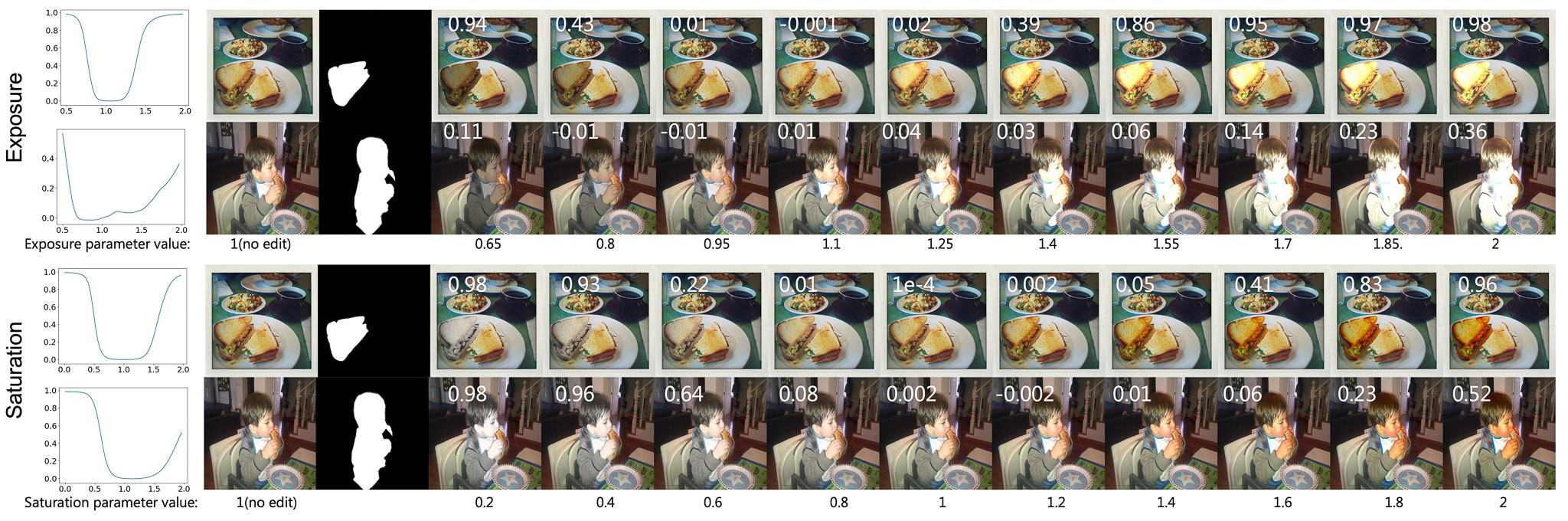}
    }
    \vspace{-0.65cm}
    \caption{The efficacy of the realism estimation network is illustrated over a range of exposure and saturation adjustments. Left is $\Delta{\mathcal{R}}$ plotted (vertical axis) for example images (second column) when selected regions (third column) are edited. Right, the edited images are shown with the corresponding change in estimated realism (inset numbers), and the value of the editing parameter applied (underneath).}
    \vspace{-0.3cm}
    \label{fig:realismnetperformance}
\end{figure*}

%% file: tex/3-realismNetwork.tex
When editing specific regions in an image, it is challenging to maintain the overall realism of the photograph.
How quickly realism starts to degrade depends on the contents and size of the image regions, the overall composition of the scene, as well as the type of edits being applied. 
This makes the problem of defining precisely when an image edit stops looking realistic particularly challenging. 

In this work, we propose to train a realism network using only realistic and unrealistic examples at the extremes. 
We generate realistic edits by slightly perturbing image values, and unrealistic edits by applying aggressive edits. 
We show that, despite being trained on binary data, our network can estimate continuous realism scores that can adapt to different types of image regions and scenes. 
Our approach was inspired by the work of Zhu et al.\cite{zhu2015learning}, who similarly learn their realism from binary real and synthetic composites.



To generate \emph{real} and \emph{fake} samples, we exploit different parameter ranges for commonly used editing operations  -- exposure, saturation, color curve, and white balancing (formal definitions in the Supplementary Material). 
For instance, increasing the exposure of a region too much can result in an unrealistic image, while a subtle increase to saturation will not signficantly affect the realism.
Based on experimentation, we determined the set of parameter ranges in Tab.~\ref{tab:editranges} to apply to image regions to create our training data. 

To generate a training example, we first select a random number of edits (between 1-4), then an order for the edit operations (e.g., exposure, saturation, color curve, white balancing), and values for each of the operations, sampled uniformly at random from the pre-specified ranges in Tab.~\ref{tab:editranges}. We apply these edits in the selected order to a region segment in an MS-COCO image~\cite{lin2014microsoft}.
Fake examples are generated by purposefully selecting extreme values. Real examples are generated by sampling subtle edits within narrower ranges.
Because of the semantic importance of human faces and increased sensitivity to edits in facial regions, we enforce smaller parameter ranges when applying edits to faces.
Fig.~\ref{fig:realismdataset} shows several examples.

We use the Pix2Pix~\cite{isola2017image} network architecture followed by two MLP layers to estimate the realism score $\mathcal{R}$ of the input. 
For our samples in the training data, $\mathcal{R}$ is defined as 1 for real and 0 for fake samples. 
We also condition the output on the input region by feeding the region's mask $M$ as input to the network. 
We use squared error~\cite{mao2017least} as the critic to compute the loss on the estimated value:
\begin{equation}
\mathcal{L}_\text{disc} =  \frac{1}{2}\mathcal{R}(I_{fake},M)^2 + \frac{1}{2}(\mathcal{R}(I_{real},M)-1)^2
\end{equation}
where $I_{fake}$ and $I_{real}$ are the generated fake and real samples. 
To measure the effect of the edit on the realism of the image, we compute the difference between the scores estimated for the original image $I$ and the edited image $I'$:
\begin{equation}
    {\Delta}\mathcal{R}(I',I,M) = \mathcal{R}(I',M) - \mathcal{R}(I,M),
\end{equation}
where the edited region is defined by the mask $M$.

\input{figures/realismdataset}

As Fig.~\ref{fig:realismnetperformance} demonstrates ${\Delta}\mathcal{R}$ gives us continuous realism values for a range of edit parameters, despite the network being trained only on extreme cases. It also shows that the range of edits that are considered realistic by our network is not the same for each image and depends on the subject and editing operation. 
We show more examples of edits that are classified realistic or unrealistic by our network in Fig.~\ref{fig:optdivedit} and the Supplementary Material.

%% file: figures/realismdataset.tex
\begin{figure}[t]
    \centering
    \footnotesize
    \resizebox{\linewidth}{!}{
    \begin{tabular}
    {  
        K{0.25\linewidth}
        K{0.25\linewidth}
        K{0.25\linewidth}
        K{0.25\linewidth}
    }
    input image & mask & real sample & fake sample
    \end{tabular}
    }
    \showimagew[\linewidth]{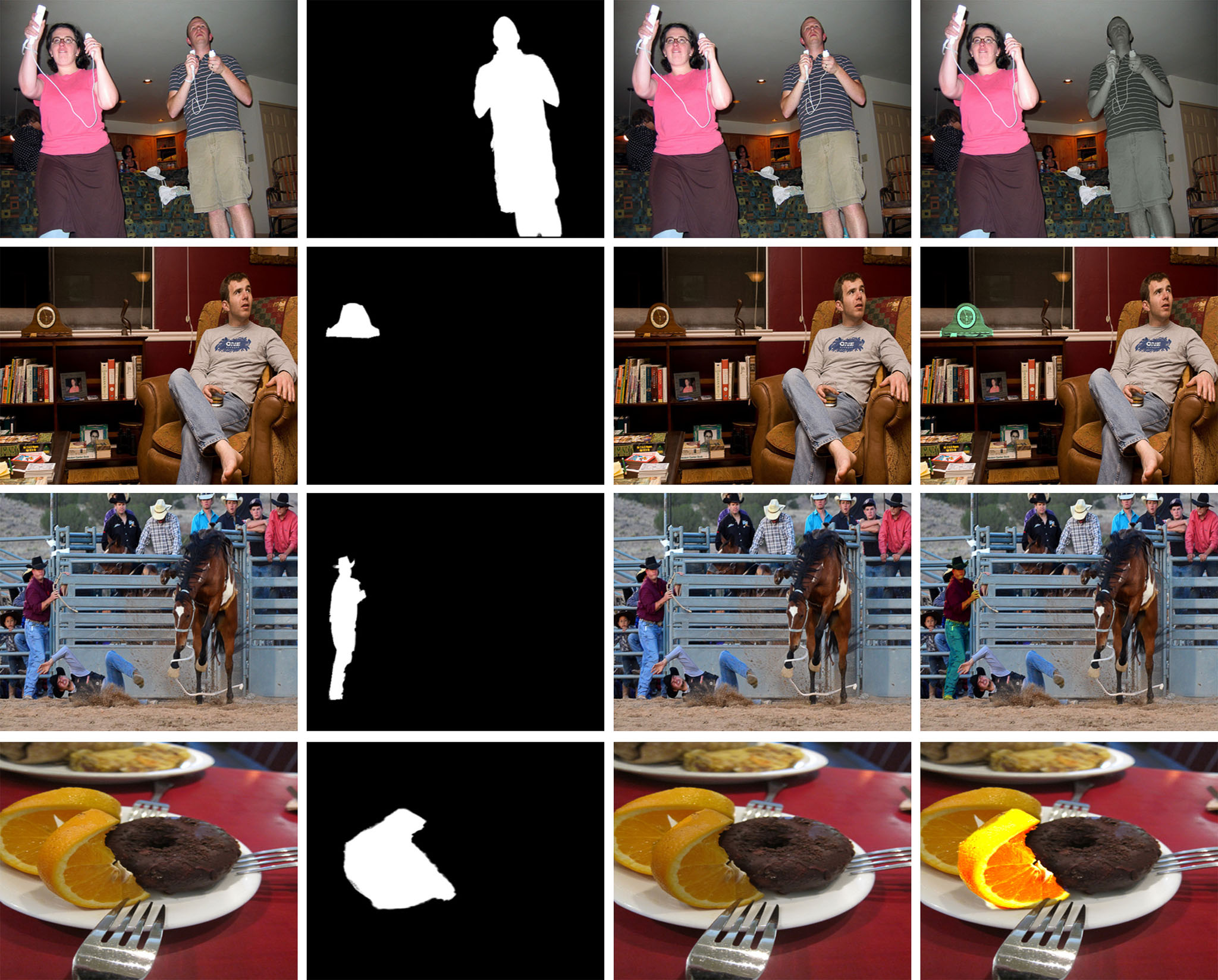}
    \vspace{-0.55cm}
    \caption{Examples of fake and real images that are used to train the realism estimation network. See Section~\ref{sec:realismmetric} for more details.}
    \vspace{-0.5cm}
    \label{fig:realismdataset}
\end{figure}

%% file: tex/4-method.tex
We develop a saliency-guided image editing pipeline that enforces our realism loss to generate realistic and effective object enhancement or distractor suppression results for a given mask. 
Our system can estimate a set of editing parameters for any permutation of 4 editing operators: exposure, saturation, color curve, and white balancing.

In constructing our system, we borrow many ideas from the existing saliency-guided image editing literature, and focus our design improvements on improving the realism of the results, especially by including our proposed realism loss. 
Since these edit operations are non-linear, different orderings of edits changes the end results.
As a result, we condition the regressed parameters on the permutation of the edit operations by feeding the permutation as an input to the network. 
More details on the architecture of the network and the embedding used to encode the permutation is included in the Supplementary Material.

\paragraph{Saliency Loss} A pretrained saliency model~\cite{JIA20EML} (SalNet) is used as a proxy for the viewer attention that would be captured by image regions before and after applying the edits, to supervise the image editing process.

We measure the change in the saliency of the region of interest as the expected value of its relative change within the masked region:
\begin{equation}
    S(I, I', M) = \mathbb{E}_{M}\left[\frac{SalNet(I) - SalNet(I')}{SalNet(I)}\right]
\end{equation}
where $\mathbb{E}$ denotes the expectation and $M$ is the region mask.

As Fig.~\ref{fig:salmodel} shows, the predicted saliency heatmaps can change drastically when applied to unrealistic edits.
As a result, relying on conventional metrics (e.g., absolute and relative differences by~\cite{gazeshift, aberman2021deep}, Binary cross entropy by~\cite{chen2019guide} and KL-divergence by~\cite{Gatys2017human}) to measure the change in saliency can cause large rewards or penalties during optimization. 
Infinitely large rewards for an unreal edit reduces the effectiveness of the realism term in the final loss function. To tackle this issue we define our saliency loss function as:

\begin{equation}
    \mathcal{L}_\text{sal} = \exp\left(w_\text{sal} S(I, I', M) \right)
\end{equation}

When saliency moves in the desired direction, the exponential squashes the loss, converging to the minimum and reducing the penalty quickly, acting as a soft margin.
This converging behaviour prevents large rewards that can be generated by unrealistic edits during training. 
The exponential term imposes larger penalties when saliency moves in the wrong direction, providing robustness against outliers and faster convergence.
$w_{sal}$ controls the absolute value of the loss to balance the weight of saliency loss in our final loss, which we set to 5 and -1 for amplification and attenuation, respectively.

\paragraph{Realism Loss} The realism loss is defined as:
\begin{equation}
 \mathcal{L}_\text{realism} = ReLU(-{\Delta}\mathcal{R}(I',I,M) - b_{r})
\end{equation}

This loss is designed to penalize unrealistic edits, while giving no rewards for edits that improve the estimated realism score of the input. This prevents the network from being penalized by images that receive a low realism score even before any edits are applied. 
ReLU and offset $b_r$ provide a margin that allows a slight decrease in realism without a penalty which we set to 0.1 in our experiments. 

We train two separate networks for each. The final network objective is the product of the two loss functions:
\begin{equation}
    \mathcal{L} = (1+\mathcal{L}_\text{realism}) \times \mathcal{L}_\text{sal}.
\end{equation}
%
In this formulation, the realism score acts as a weight for the penalty imposed on the change in the saliency. This allows us to balance the realism and saliency objectives. 

We use an EfficientNet-lite3~\cite{tan2019efficientnet} backbone and cascaded MLP layers as decoders to estimate parameters for each of the edit operations. 
A detailed explanation of the architecture specifics, datasets and training is provided in Supplementary Material.

%% file: tex/5-experiments.tex
\input{figures/deepsalcomp}
\input{figures/lookherecomp}

We compare our method against state of the art saliency based image editing approaches -- Deepsal~\cite{aberman2021deep}, Gazeshift~\cite{gazeshift} and MEC~\cite{mechrez2019saliency}. MEC provides results on their dataset alongside pre-computed results of WRS~\cite{wong2011saliency} and OHR~\cite{Mateescu2014attention} on the same dataset.
We use this dataset to compare against WRS and OHR as well as MEC. \footnote{Deepsal, WRS and MEC do not provide an open-source implementation. Hence, we relied on the results included on their project pages. Also, Deepsal authors kindly provided us with results on Adobe Stock dataset for their ``convolutional network" variation.}

The EfficientNet~\cite{tan2019efficientnet} backbone used in our architecture is known for its small size.
Our results are thus generated significantly faster than the other state-of-the-art (SOTA) methods with bulkier architectures and slower per-image optimizations.
Based on speed measures reported in ~\cite{gazeshift} Table 1c, MEC takes more than a day, OHR needs 30 seconds and Gazeshift takes 8 seconds to process each image, while our model requires only 8ms per image. 

We present both qualitative and quantitative results. Since our method takes the permutation of the edits as input during inference time we select the permutation at random for the presented results unless mentioned otherwise.

\subsection{Qualitative Comparison}
Figs.~\ref{fig:deepsalcomp},~\ref{fig:lookherecomp}, and~\ref{fig:mechrez} illustrate our results compared to the SOTA.
They show our method performs different edits based on the contents of the image.
It can apply more significant color changes that camouflage the distractor (2\ts{nd} and 4\ts{th} rows of Fig.~\ref{fig:lookherecomp}, 3\ts{rd} row of Fig.~\ref{fig:deepsalcomp}) or very subtle edits for human faces (1\ts{st} row of Fig.~\ref{fig:lookherecomp}). The intensity and characteristics of the applied edits depends on semantics.

The use of adversarial loss in Gazeshift~\cite{gazeshift} and the regularization used in Deepsal~\cite{aberman2021deep} constrain the edits their methods apply without taking realism explicitly into account. 
As results show, they often apply unrealistic edits (e.g., the camouflaged signs in Fig.~\ref{fig:deepsalcomp} or the unattural skin tone and the color artifacts in Fig.~\ref{fig:lookherecomp}) or very subtle edits with lower effectiveness. 

MEC~\cite{mechrez2019saliency} reuses the color patterns and textures available in the image to update the target region. On the other hand different regions and textures can correspond to different semantics. Consequently, as illustrated in Fig.~\ref{fig:mec_amply} this method can apply incompatible color and texture values to produce unrealistic edits (green crocodile eye, orange traffic sign) or ineffective enhancements (brown bird). Fig.~\ref{fig:mec_atten} provides a comparison on their \textit{distractor suppression} image set. Our method performs comparable in terms of effectiveness and generates realistic results consistently.

OHR~\cite{Mateescu2014attention} tries to maximize the color distinction between the masked region and the rest of the image for the image enhancement task. 
Without explicit realism modeling, it tends to generate unrealistic colors (e.g., blue crocodile, bird, and horse in Fig.~\ref{fig:mec_amply}). While incorrect colors increase the saliency of these regions, they do so at the cost of realism. For similar reasons, this method is ineffective suppressing distractors (Fig.~\ref{fig:mec_atten}).

WRS~\cite{wong2011saliency} does not generate unrealistic images, but also makes edits that are hardly noticeable, and less effective at enhancing or suppressing the target regions.
We believe this is due to the purposely limited range of allowed edit parameters (luminance, saturation and sharpness). 

\input{figures/mechrez}

\subsection{What Do Photographers Think?}
To include the perspective of professional photographers in comparing our results to others, we ran a user study.
We report our results using three choices of parameter orderings: choosing the one that achieves the \textit{Best Saliency}, the one that generates the \textit{Best Realism} (according to our realism estimation network), and a permutation of parameters selected at \textit{Random} as used for the qualitative figures.

\paragraph{User Study} 
We recruited 8 professionals from UpWork, all of whom have multiple years of experience with photography, as well as using Photoshop to edit photos.
We used the Appen platform for hosting our rating tasks. 

Our study participants were presented with a panel of 3 images: the original image, mask, and an edited result from one of methods evaluated.
They were asked to \textit{``rate each image based on 2 criteria"} - effectiveness and realism, with the following definitions provided for the \textit{attenuate} version of the task: \textit{``The images were edited to make certain objects and regions less distracting.
An image edit is effective if the masked objects/regions have indeed become less distracting. 
An image edit is realistic if the photo does not look edited."} For the \textit{amplify} version of the task, the wording for effectiveness was modified to: \textit{``The images were edited to make certain objects and regions pop-out (more attention-capturing, or salient).
An image edit is effective if the masked objects/regions have indeed become more attention-capturing.''}
Images were randomly shuffled in each task, so the photographers rated the images and methods independently of each other.

\input{tables/userstudy_adobestock}

\input{tables/userstudy_mechrez}

\paragraph{Results} 
In Tab.~\ref{tab:userstudyresults} we compare our approach to GazeShift and Deepsal on the 30 Adobe Stock images from~\cite{gazeshift}.
We find that our approach achieves significantly higher scores for both effectiveness and realism compared to GazeShift in the attenuation task. This matches our qualitative observations that GazeShift is not successful at the task of attenuating distractor. GazeShift specializes in amplifying saliency in image regions, and we achieve similar performance on this task, while also maintaining significantly higher realism levels. 
In addition, results show a poor effectiveness score for Deepsal as a result of subtle edits in Fig.~\ref{fig:lookherecomp}. Subtle edits mean the realism score remains high since the results are almost identical to the original images.

Since Deepsal was ineffective on Adobe Stock images, to provide a fair comparison we also compare to Deepsal on 14 images they provided on their project page in Tab.~\ref{tab:userstudydeepsal}.
We achieve significantly higher realism scores while being similarly effective at reducing the saliency of the distractors.
This matches our qualitative observations that Deepsal edits can be quite extreme and not always photo realistic.

Tab.~\ref{tab:userstudymec} shows user study results on Mechrez dataset~\cite{mechrez2019saliency}.\footnote{Dataset has only 10 images for attenuation task, which is inadequate for a meaningful user study. Hence we only provide amplification results.}
We used 77 images from the dataset to perform the user study. Results confirm that our results are superior in the realism while we achieve a comparable effectiveness compared MEC. WRS's low effectiveness yields a high realism score as its results are almost identical to the input; while the unrealistic color changes by OHR result in low realism and effectiveness scores. 


\subsection{Ablation Study}
We trained a variation of our method in which instead of a fixed realism score estimation model we used a discriminator as adversarial loss.
We trained the discriminator as part of an adversarial training approach, similar to related work~\cite{gazeshift, chen2019guide, Gatys2017human}. We used the input image as the real sample and the generated image as the fake sample during training.
Fig.~\ref{fig:ablation} illustrates sample results with this training strategy. 
Since the discriminator is trained to penalize "any edits" applied in the previous steps of training it encourages the network to apply subtle edits and hence a drop in effectiveness of the method. On the other hand, due to the lack of explicit realism training, the edits are unrealistic while the effectiveness is reasonable.
Ratings reported in Tab.~\ref{tab:ablation} also confirm our findings.

\input{figures/ablation}
\input{tables/ablation}

\input{figures/diveristy}

\subsection{Diversity and Optimality of Estimated Edits} Fig.~\ref{fig:diveristy} illustrates the distribution of edit parameters estimated by our parameter estimation network for different images on ADE20K~\cite{zhou2019semantic,zhou2017scene} dataset. It shows that edit parameters are different for each image and is based on its content. Also, it shows that the range of estimated edits is not the same as the ranges used in Tab.~\ref{tab:editranges} for real samples.

To evaluate if the estimated edits are close to optimal with respect to realism, we provide Fig.~\ref{fig:optedit}. In the figure we show a realism heatmap obtained by adding a small additive constant to the estimated edit parameter of \textit{saturation} and \textit{exposure}. Heatmaps shows the estimated edit parameters (center of the heatmap) are in the optimal realism region. Changing the edit parameters in each direction reduces the realism of the end result.

\subsection{Generalization to Multiple Image Regions}
Since our model only modifies the region of interest, and performs a forward pass efficiently, we can run it on multiple regions and multiple masks by generating edit parameters for each region, in an iterative manner. Examples are provided in Figs.~\ref{fig:teaser} and \ref{fig:multimask}. We used the same approach with Gazeshift~\cite{gazeshift}, which edits the whole image by estimating two sets of edit parameters, one for the region of interest (foreground) and one for the background. This formulation of Gazeshift makes iterative editing impractical, since there would be contradictory objectives between the iterations (what is foreground in one iteration becomes a background in the next iteration).
For a more practical comparison, we omit background edits when running Gazeshift. Figure~\ref{fig:multimask} shows that Gazeshift performance suffers on an iterative saliency enhancement task, but our method is able to generalize to multiple regions robustly. 
\input{figures/multimask}

\subsection{Limitations}
The global edits (applying the same edits to every pixel inside a mask), used in both our method and Gazeshift~\cite{gazeshift} require an accurate mask of the target region. As shown in Fig.~\ref{fig:limitation} mask imperfections can cause unsmooth transitions around the boundaries. In these cases, pixel-wise optimization approaches likes Deepsal~\cite{aberman2021deep} and MEC~\cite{mechrez2019saliency} do not suffer from heavy artifacts due to mask imperfections.

\input{figures/limitation}

%% file: figures/deepsalcomp.tex
\begin{figure}[t]
    \centering
    \footnotesize
    \resizebox{\linewidth}{!}{
    \begin{tabular}
    {  
        K{0.25\linewidth}
        K{0.25\linewidth}
        K{0.25\linewidth}
    }
    Input image & Ours & Deepsal~\cite{aberman2021deep} 
    \end{tabular}
    }
    
    \showimagew[\linewidth]{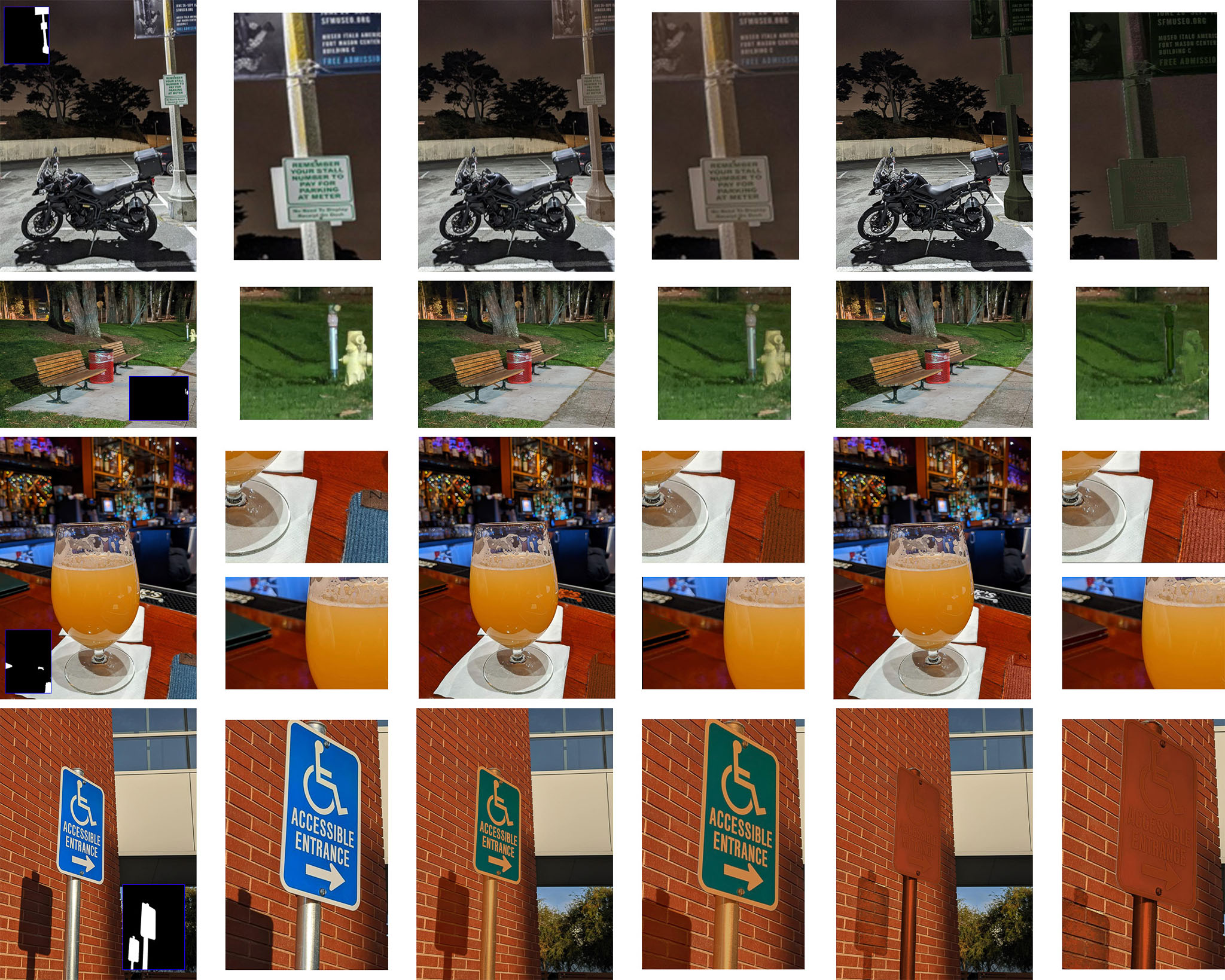}
    \vspace{-0.55cm}
    \caption{Saliency attenuation compared to Deepsal~\cite{aberman2021deep} on the images provided by the authors on their project webpage. Our method is able to effectively attenuate the saliency of the target region without applying an unrealistic camouflage.}
    \vspace{-0.6cm}
    \label{fig:deepsalcomp}
\end{figure}

%% file: figures/lookherecomp.tex
\begin{figure*}[tb]
    \centering
    
    \footnotesize
    \resizebox{\linewidth}{!}{
    \begin{tabular}
    {
        K{0.17\linewidth}
        K{0.5\linewidth}
        K{0.33\linewidth}
    }
      & Attenuation & Amplification
    \end{tabular}
    }
    
    \footnotesize
    \resizebox{\linewidth}{!}{
    \begin{tabular}
    {
        K{0.166\linewidth}
        K{0.166\linewidth}
        K{0.166\linewidth}
        K{0.166\linewidth}
        K{0.166\linewidth}
        K{0.166\linewidth}
    }
    Input image & Ours & Gazeshift~\cite{gazeshift} & Deepsal~\cite{aberman2021deep}&  Ours & gazeshift~\cite{gazeshift}
    \end{tabular}
    }
    \showimagew[\linewidth]{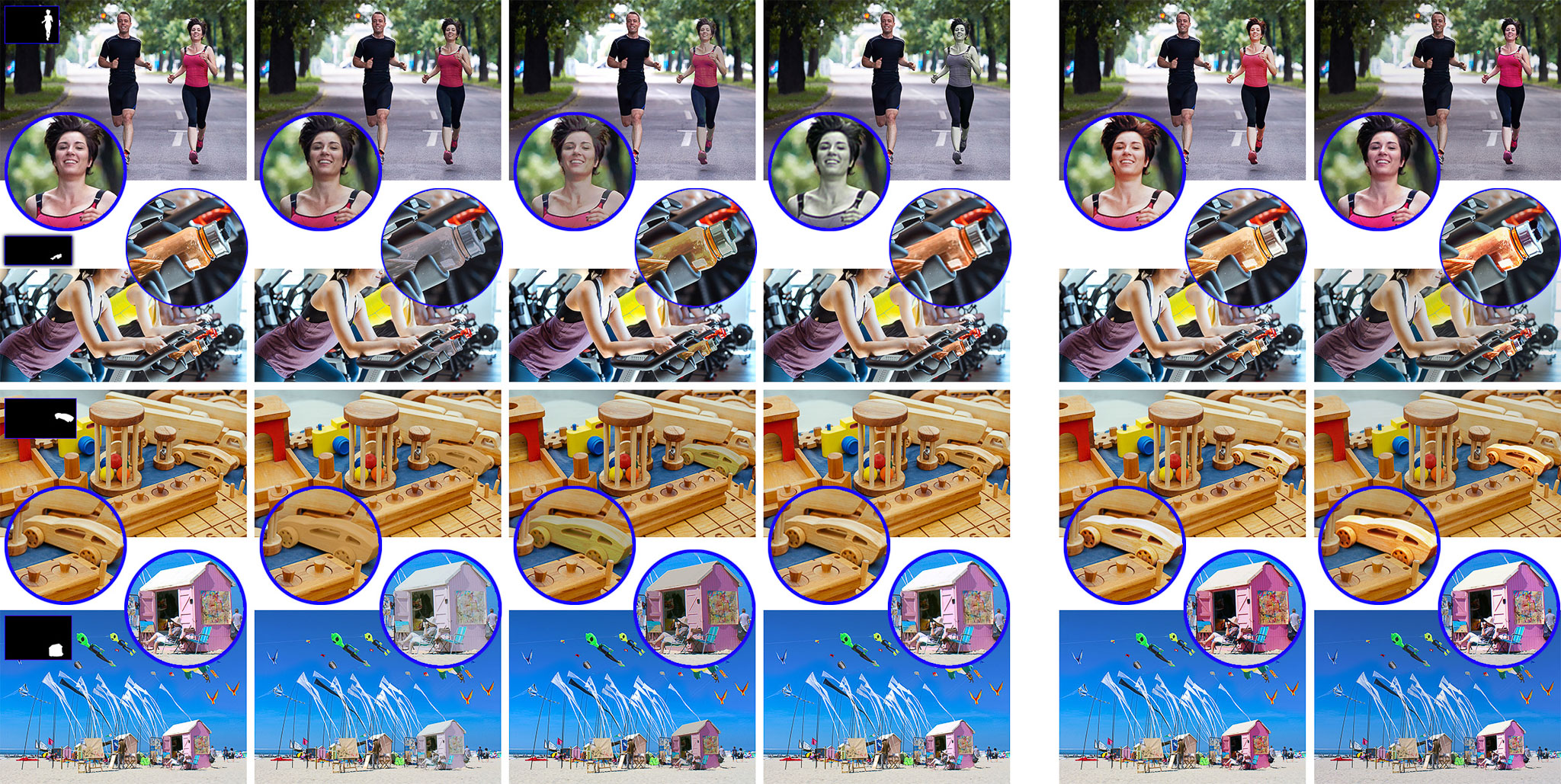}
    \vspace{-0.55cm}
    \caption{Saliency modulation compared to GazeShift~\cite{gazeshift} and Deepsal~\cite{aberman2021deep} on Adobe Stock images from~\cite{gazeshift}.}
    \label{fig:lookherecomp}
\end{figure*}

%% file: figures/mechrez.tex
\begin{figure*}[tb]
    \centering
    
    \begin{subfigure}[t]{0.49\linewidth}
    \footnotesize
    \resizebox{\linewidth}{!}{
    \begin{tabular}
    {   
        K{0.2\linewidth}
        K{0.2\linewidth}
        K{0.2\linewidth}
        K{0.2\linewidth}
        K{0.2\linewidth}
    }
    Input image & OHR~\cite{Mateescu2014attention} & WRS~\cite{wong2011saliency} & Mechrez~\cite{mechrez2019saliency} & Ours
    \end{tabular}
    }
    \showimagew[\linewidth]{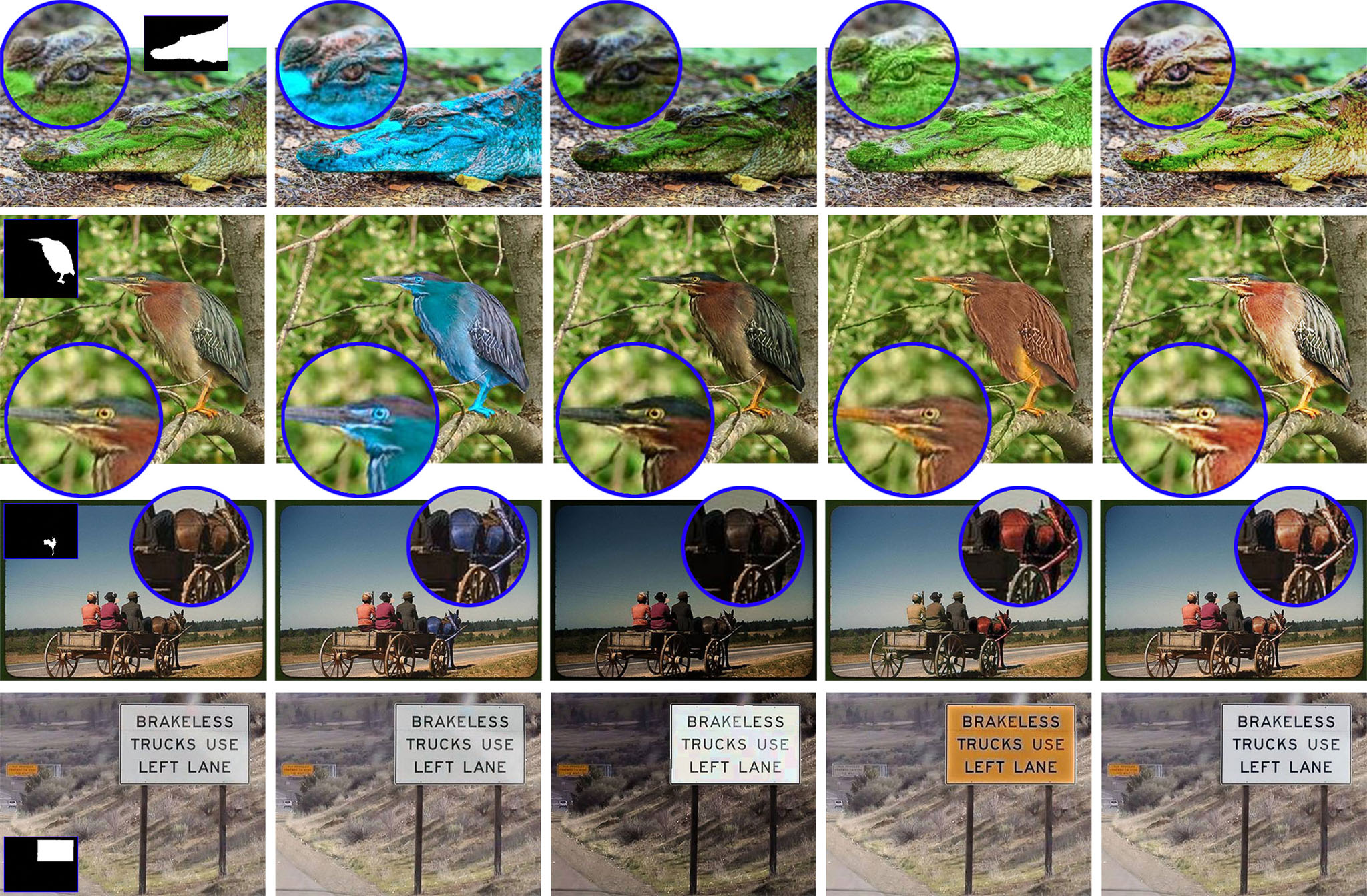}
    \caption{Image Enhancement (Amplification)}
    \label{fig:mec_amply}
    \end{subfigure}
    \hfill
    \begin{subfigure}[t]{0.49\linewidth}
    \footnotesize
    \resizebox{\linewidth}{!}{
    \begin{tabular}
    {   
        K{0.2\linewidth}
        K{0.2\linewidth}
        K{0.2\linewidth}
        K{0.2\linewidth}
        K{0.2\linewidth}
    }
    Input image & OHR~\cite{Mateescu2014attention} & WRS~\cite{wong2011saliency} & Mechrez~\cite{mechrez2019saliency} & Ours
    \end{tabular}
    }
    \showimagew[\linewidth]{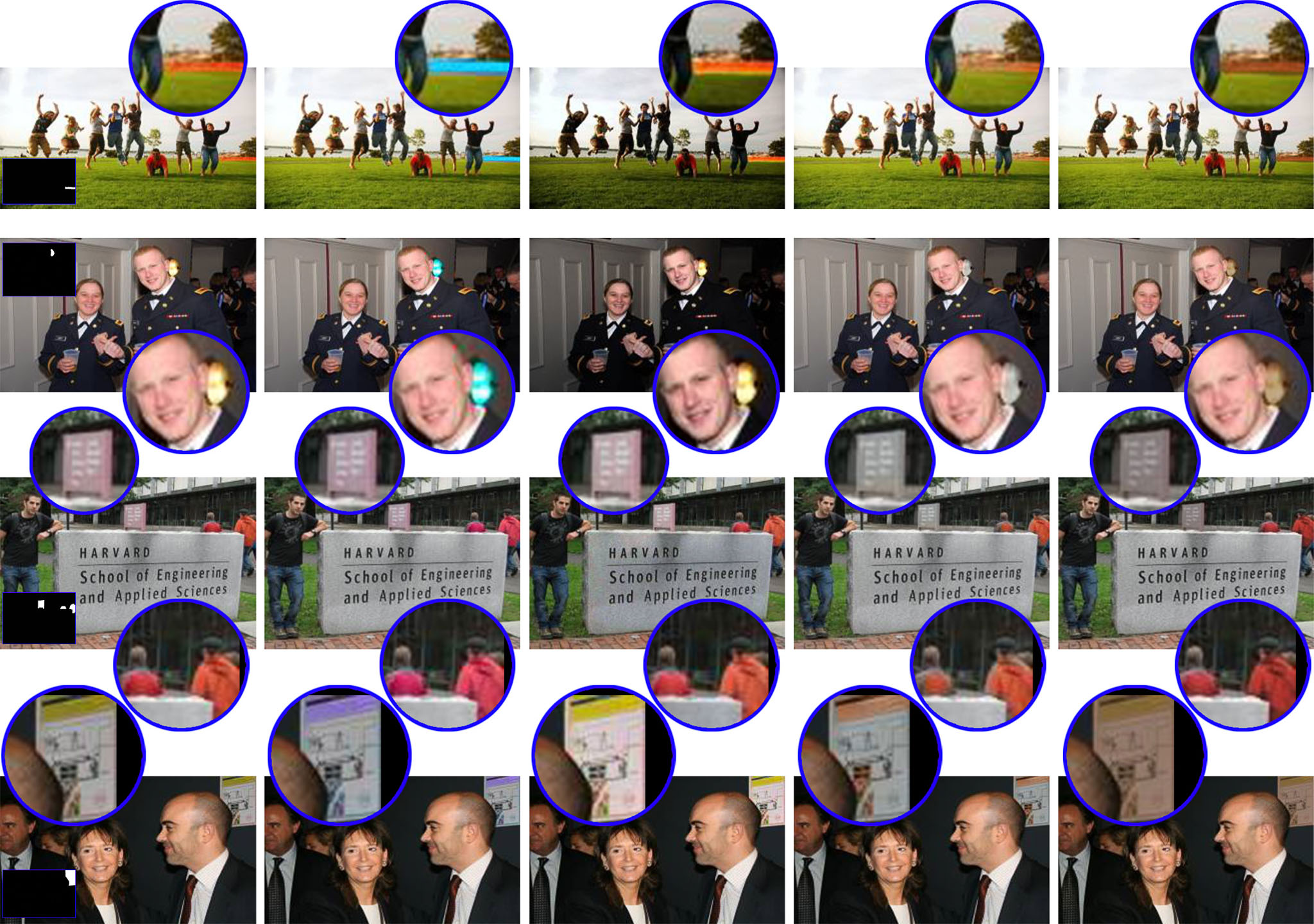}
    \caption{Distractor Suppression (Attenuation)}
    \label{fig:mec_atten}
    \end{subfigure}
    \vspace{-0.25cm}
    \caption{Saliency modulation compared to MEC~\cite{mechrez2019saliency}, WRS~\cite{wong2011saliency} and OHR~\cite{Mateescu2014attention} on the Mechrez dataset~\cite{mechrez2019saliency}.}
    \vspace{-0.5cm}
    \label{fig:mechrez}
\end{figure*}

%% file: tables/userstudy_adobestock.tex
\begin{table}[tb]
\caption{Photographer ratings (on a 1 to 10 scale, higher is better) of effectiveness (i.e., achieve objective of attenuation or amplification of saliency) and realism (i.e., photo looks natural) on the dataset of 30 Adobe stock images. Numbers are the mean score across 8 photographers, with standard deviation in parentheses.}
\vspace{-0.25cm}
\resizebox{\linewidth}{!}
{
\begin{tabular}{l|ll|ll}

& \multicolumn{2}{c|}{Saliency Attenuation} & \multicolumn{2}{c}{Saliency Amplification}\\
Method & Effectiveness $\uparrow$ & Realism $\uparrow$ & Effectiveness $\uparrow$ & Realism $\uparrow$ \\
\hline
GazeShift~\cite{gazeshift}                   & 4.78 (2.89)  & 5.93 (3.13) & 7.36 (2.37) & 7.07 (2.76) \\     
DeepSal~\cite{aberman2021deep}               & 4.04 (2.90) & 8.49 (2.72) & - & - \\
\hline
Ours - Best Realism        & 6.56 (2.73)  & 6.78 (2.70)  & 7.39 (2.17) & 8.31 (1.89) \\  
Ours - Random               & 6.36 (2.79)  & 6.34 (2.88) & 7.36 (2.21) & 8.27 (1.94)  \\         
Ours - Best Saliency       & 6.64 (2.79) & 6.31 (2.70)  & 7.50 (2.08) & 8.15 (2.10)  \\     
        
\end{tabular}
}


\label{tab:userstudyresults}
\end{table}

%% file: tables/userstudy_mechrez.tex

\begin{table}[tb]
\caption{Photographer ratings as in Tab.~\ref{tab:userstudyresults} on (a) Mechrez~\cite{mechrez2019saliency} dataset and (b) the 14 images from DeepSal project webpage~\cite{aberman2021deep} }
\vspace{-0.25cm}

\begin{subtable}[t]{0.49\linewidth}
\resizebox{\linewidth}{!}
{
\begin{tabular}{l|ll}

& \multicolumn{2}{c}{Saliency Attenuation} \\
Method & Effectiveness $\uparrow$ & Realism $\uparrow$ \\
\hline
Deepsal~\cite{aberman2021deep}                 & 7.08 (2.84)  & 5.82 (3.43) \\     
\hline
Ours - Random               & 6.83 (2.52)  & 7.41 (2.70)  \\         
\end{tabular}
}
\caption{}
\label{tab:userstudydeepsal}
\end{subtable}
\hfill
\begin{subtable}[t]{0.49\linewidth}
\resizebox{\linewidth}{!}
{
\begin{tabular}{l|ll}

&  \multicolumn{2}{c}{Saliency Amplification}\\
Method &  Effectiveness $\uparrow$ & Realism $\uparrow$ \\
\hline
MEC~\cite{mechrez2019saliency}                          & 7.06 (2.68) & 7.31 (2.93) \\   
WRS~\cite{wong2011saliency}                             & 5.41 (3.22) & 7.97 (2.70) \\
OHR~\cite{Mateescu2014attention}                        & 7.04 (3.04) & 5.18 (3.76) \\
\hline
Ours - Random                                           & 6.24 (2.9) & 8.88 (1.74)  \\   
            
\end{tabular}
}
\caption{}
\label{tab:userstudymec}
\end{subtable}

\vspace{-0.6cm}
\label{tab:userstudydeepsalandmec}
\end{table}

%% file: figures/ablation.tex
\begin{figure}[tb]
    \centering
    \footnotesize
    \resizebox{\linewidth}{!}{
    \begin{tabular}
    {  
        K{0.33\linewidth}
        K{0.33\linewidth}
        K{0.33\linewidth}
    }
    Input image & Ours & Adversarial training
    \end{tabular}
    }
    
    \showimagew[\linewidth]{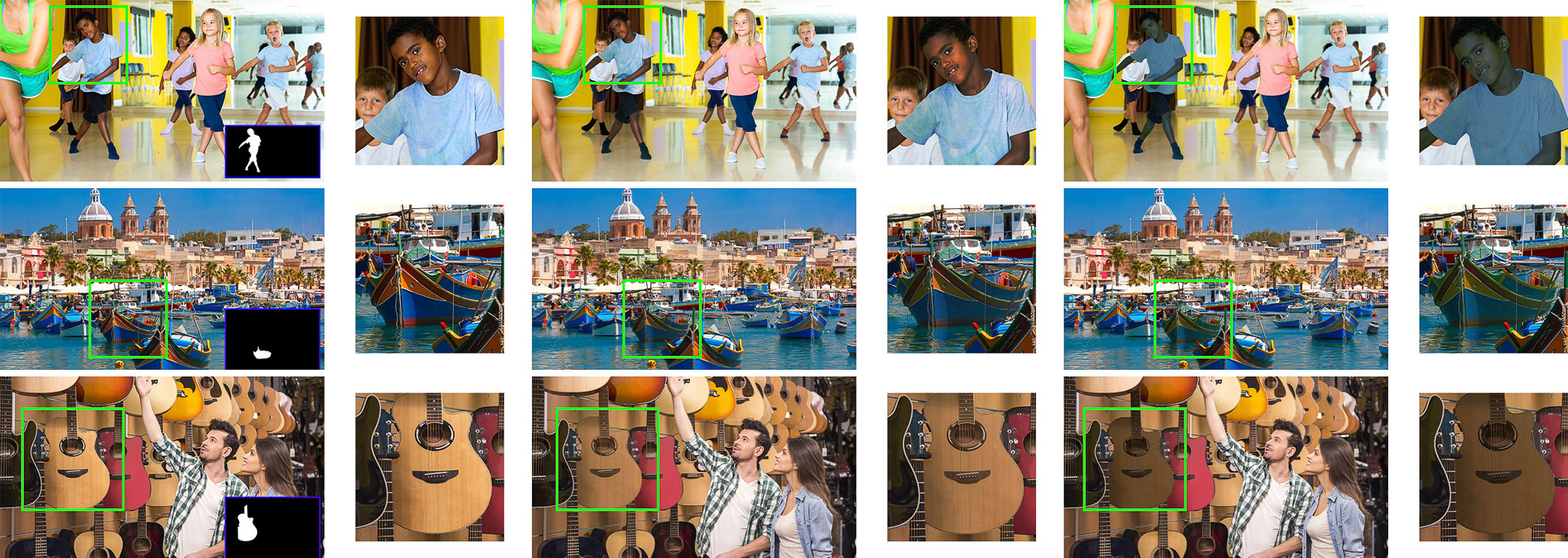}
    \vspace{-0.55cm}
    \caption{When the model trained via adversarial training produces results that are effective at reducing saliency, the resulting images are not realistic according to our user study.}
    \label{fig:ablation}
\end{figure}

%% file: tables/ablation.tex
\begin{table}[tb]
\centering
\caption{Photographer ratings as in Tab.~\ref{tab:userstudyresults} comparing our main method to a variation with adversarial training instead of our fixed realism network.}
\vspace{-0.25cm}
\resizebox{0.7\linewidth}{!}
{
\begin{tabular}{l|ll}
& \multicolumn{2}{c}{Saliency Attenuation} \\
Method & Effectiveness $\uparrow$ & Realism $\uparrow$ \\
\hline
Adversarial Training         & 5.06 (2.84)  & 7.36 (3.07) \\     
Ours - Random               & 6.36 (2.79)  & 6.34 (2.88) \\         
\end{tabular}
}
\label{tab:ablation}
\end{table}

%% file: figures/diveristy.tex
\definecolor{myred}{RGB}{132, 0, 0}
\definecolor{myblue}{RGB}{0, 0, 132}

\begin{figure*}[!ht]
    \centering
    \begin{subfigure}[t]{0.58\linewidth}
    \showimagew[\linewidth]{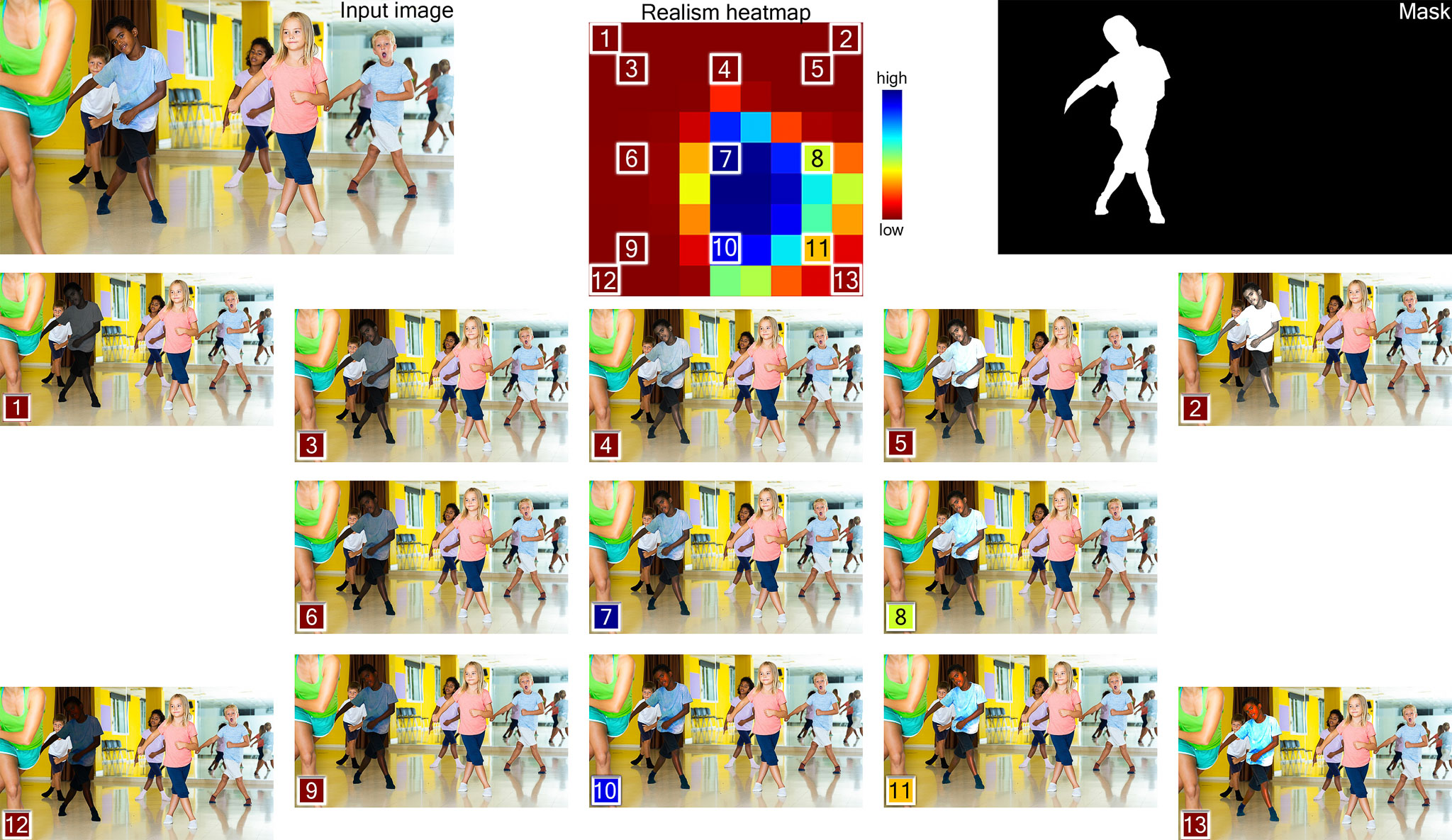}
    \caption{A heatmap visualizes the realism score achieved when we change the estimated saturation (x-axis) and exposure (y-axis). Our estimated values (center of the heatmap) achieve the optimal realism while changing the parameters in any direction reduces the realism. Sample edited images and their corresponding location in the heatmap are also visualized.}
    \label{fig:optedit}
    \end{subfigure}
    \hfill
    \begin{subfigure}[t]{0.38\linewidth}
    \showimagew[\linewidth]{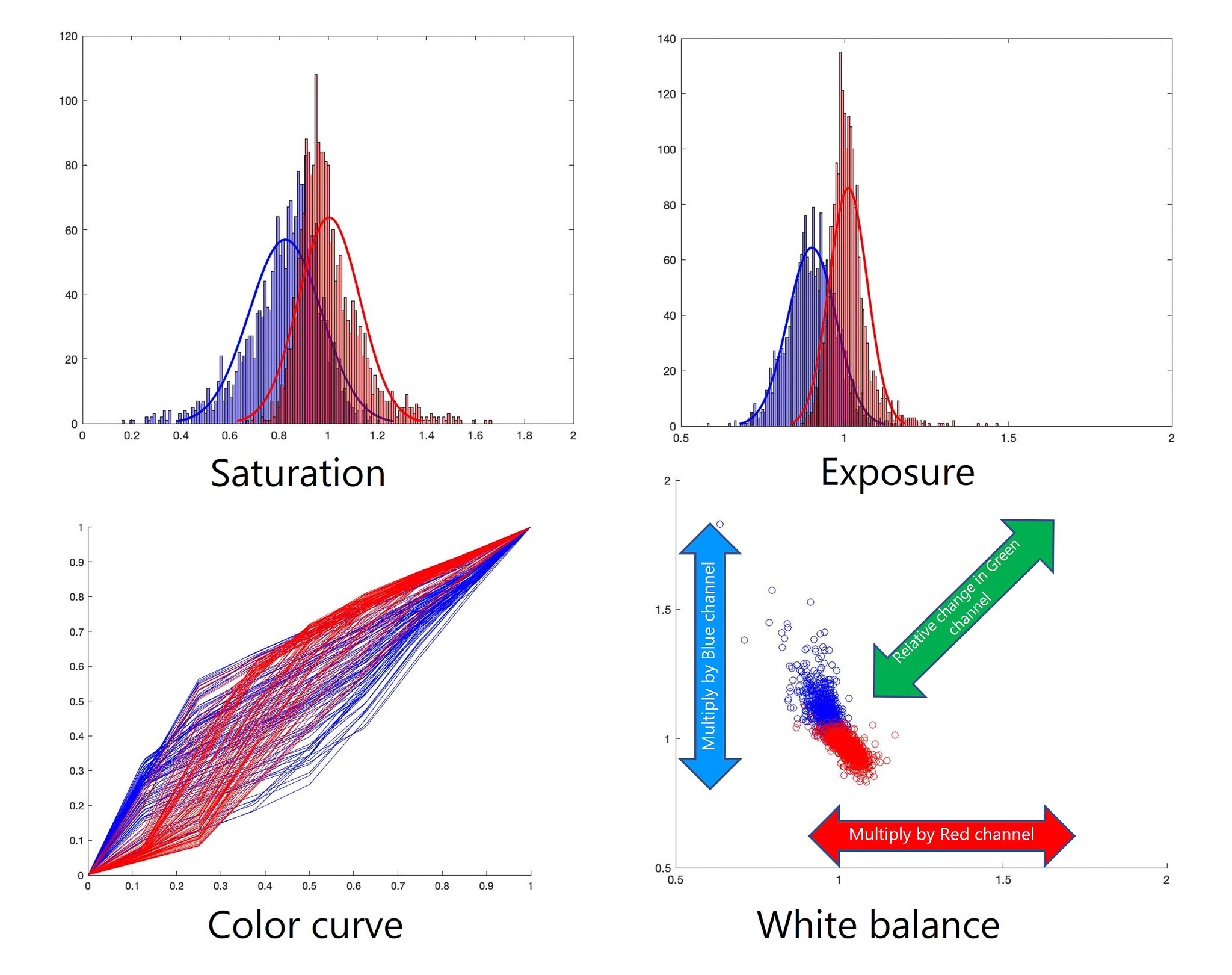}
    \caption{The diversity of estimated parameters on ADE20K~\cite{zhou2019semantic,zhou2017scene} dataset. The x-axis is the range of each parameter. The attenuation task is blue, and amplification is labeled red.}
    \label{fig:diveristy}
    \end{subfigure}
    \vspace{-0.15cm}
    \caption{Visualizing diversity and optimality of edit pararmeters estimated by our method}
    \vspace{-0.4cm}
    \label{fig:optdivedit}
\end{figure*}

%% file: figures/multimask.tex
\begin{figure}[t]
\centering
\showimagew[\linewidth]{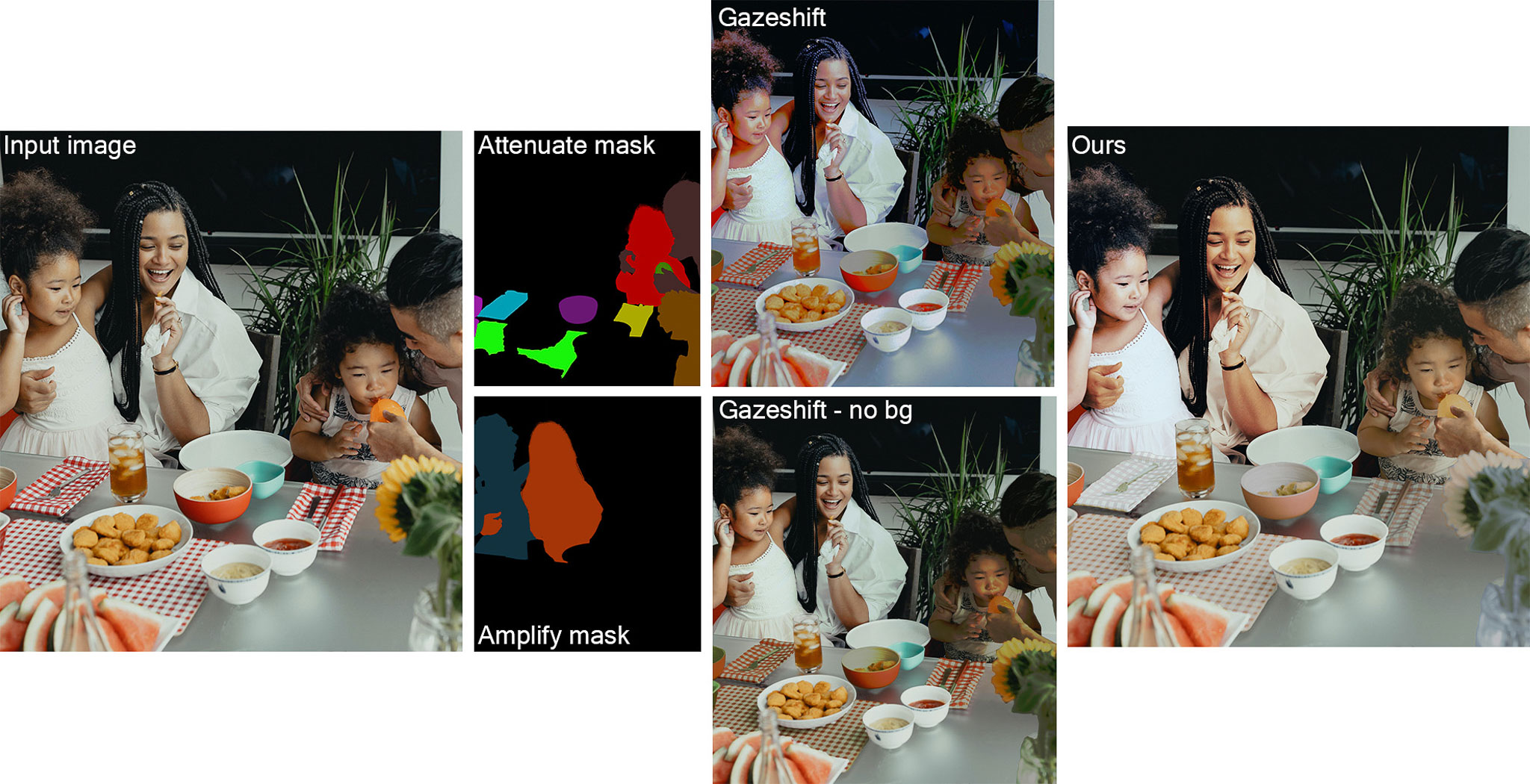}
\vspace{-0.5cm}
\caption{
Given an input image and masks to attenuate and amplify(left), Gazeshift when used iteratively on each object suffers from color artifacts (center top, faces, bowl and watermelons). Ours produces a notably more realistic and effective result (right). Contradictory objective of edits applied to background and fore-
ground, Gazeshift fails to generalize to multiple regions and omitting the background edits (center bottom) reduces the effectiveness of the edits. Image credit:
\myhref[purple]{https://unsplash.com/photos/gorbBYbo6KM}{@$\text{tysonbrand}$} }
\vspace{-0.5cm}
\label{fig:multimask}
\end{figure}

%% file: figures/limitation.tex
\begin{figure}[t]
\centering
\showimagew[\linewidth]{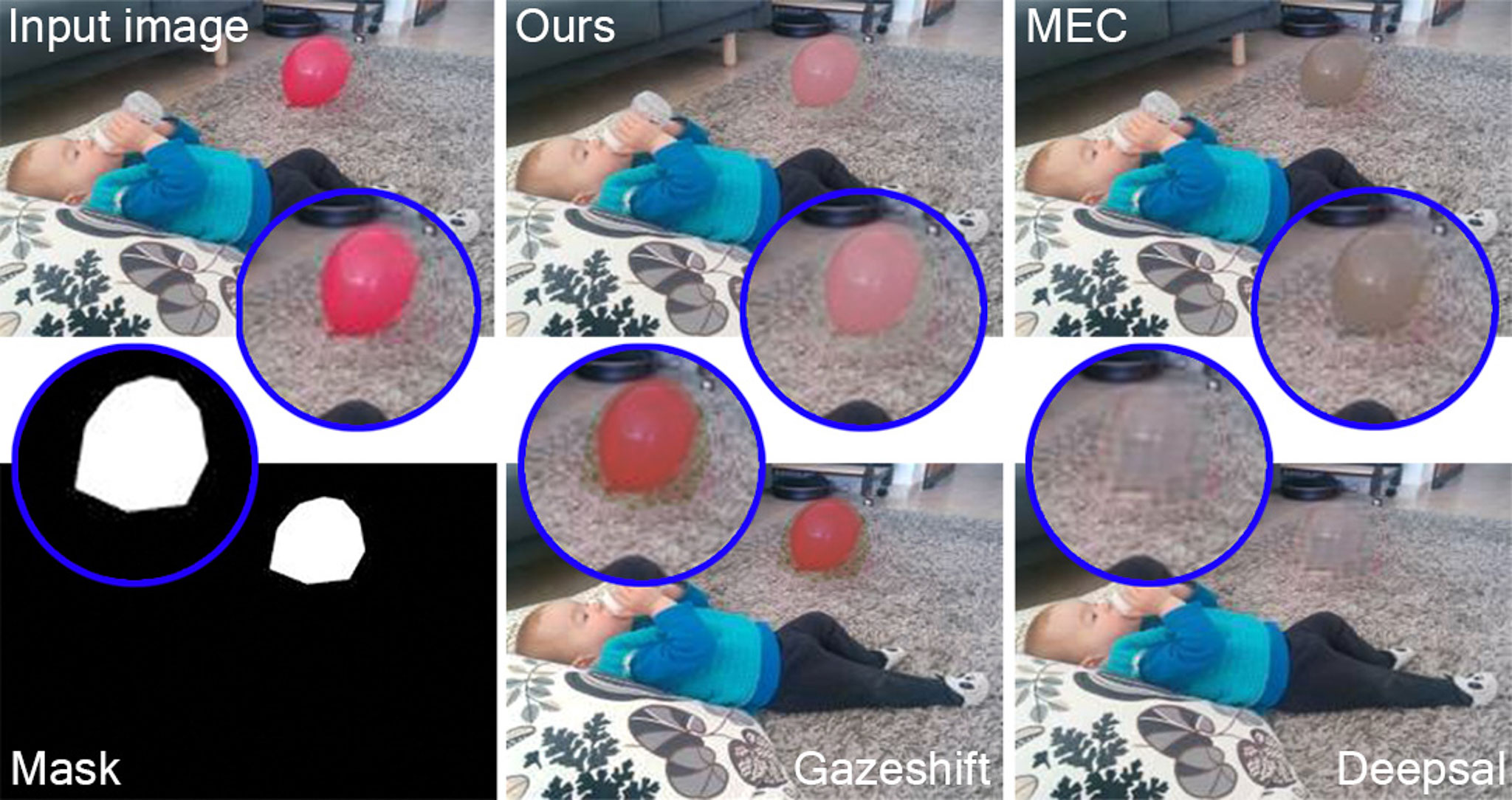}
\vspace{-0.5cm}
\caption{The effect of non-smooth mask boundaries. Left, an input image has a mask with a sharp edge.  Center, our method and Gazeshift~\cite{gazeshift} produce strong boundary artifacts around the mask region (see inset). Right, MEC~\cite{mechrez2019saliency} and Deepsal~\cite{aberman2021deep} do not exhibit this problem because they operate in a pixel-wise manner.}
\vspace{-0.5cm}
\label{fig:limitation}
\end{figure}

%% file: tex/6-conclusion.tex
We describe a method to edit images using conventional image editing operators to attenuate or amplify the attention captured by a target region while preserving image realism. Realism is achieved by introducing an explicit, and separate realism network that is pre-trained to distinguish edited images. This strategy to achieve realism is distinct from prevailing approaches, including adversarial training schemes, as it introduces an additional form of weak supervision---manually specified ranges of parameter values that correspond to realistic and unrealistic (``fake") edits. Training with this realism critic makes it possible to estimate saliency modulating image edits that are significantly more realistic and robust. Together with our millisecond-level inference time, our approach offers a practical and deployable application of saliency guided image editing. 